\documentclass[letterpaper, 10 pt, conference]{ieeeconf}  

\IEEEoverridecommandlockouts                              

\usepackage{hyperref}
\usepackage{amsmath}
\usepackage{array}
\usepackage{multirow}
\usepackage{longtable}
\usepackage{rotating}
\usepackage{makecell}
\usepackage{booktabs}
\usepackage[table]{xcolor}
\usepackage{subfig}
\usepackage{algorithm}
\usepackage{algpseudocode}
\usepackage{amssymb, amsfonts}

\hypersetup{hidelinks,
	colorlinks=true,
	allcolors=black,
	pdfstartview=Fit,
	breaklinks=true}
    
\title{\LARGE \bf
PL-VIWO: A Lightweight and Robust Point-Line Monocular Visual Inertial Wheel Odometry
}

\author{Zhixin Zhang$^{1}$, Wenzhi Bai$^{1}$, Liang Zhao$^{2}$, Pawel Ladosz$^{3}$
\thanks{This work was partially funded by the Robotics and AI Collaboration(RAICo).}
\thanks{$^{1}$Zhixin Zhang, Wenzhi Bai are with the Department of Electrical and Electronic Engineering, University of Manchester, Manchester M13 9PL, U.K.
Email: {\tt\small \{zhixin.zhang, wenzhi.bai\} @manchester.ac.uk}}
\thanks{$^{2}$Liang Zhao is with the School of Informatics, University of Edinburgh, Edinburgh EH8 9AB, U.K.
Email: {\tt\small liang.zhao@ed.ac.uk}}
\thanks{$^{3}$Pawel Ladosz is with the Department of Mechanical, Aerospace and Civil Engineering, University of Manchester, Manchester M13 9PL, U.K.
Email: {\tt\small pawel.ladosz@manchester.ac.uk}}}

\begin{document}

\maketitle 
\thispagestyle{empty}
\pagestyle{empty}

\begin{abstract}

This paper presents a novel tightly coupled Filter-based monocular visual-inertial-wheel odometry (VIWO) system for ground robots, designed to deliver accurate and robust localization in long-term complex outdoor navigation scenarios. As an external sensor, the camera enhances localization performance by introducing visual constraints. However, obtaining a sufficient number of effective visual features is often challenging, particularly in dynamic or low-texture environments. To address this issue, we incorporate the line features for additional geometric constraints. Unlike traditional approaches that treat point and line features independently, our method exploits the geometric relationships between points and lines in 2D images, enabling fast and robust line matching and triangulation. Additionally, we introduce Motion Consistency Check (MCC) to filter out potential dynamic points, ensuring the effectiveness of point feature updates. 
The proposed system was evaluated on publicly available datasets and benchmarked against state-of-the-art methods. Experimental results demonstrate superior performance in terms of accuracy, robustness, and efficiency. The source code is publicly available at: \href{https://github.com/Happy-ZZX/PL-VIWO}{https://github.com/Happy-ZZX/PL-VIWO}.

\end{abstract}

\section{Introduction}
Achieving robust and accurate localization in complex outdoor environments remains a critical challenge for ground robot technology. Visual Odometry systems which rely solely on image data, often struggle in outdoor navigation, leading to unstable and inaccurate localization. To address these limitations, integrating interoceptive sensors, such as IMUs and wheel encoders has been proven to be an effective approach. This integration not only improves system robustness \cite{c1} but also mitigates errors caused by unobservable states during degenerate motions \cite{c2}. Furthermore, Multi-sensor-fusion-based systems can provide stable estimation results based on measurements of interoceptive sensors in texture-less, dynamic, and poor illumination environments. These advantages have made Visual-Inertial-Wheel Odometry a widely adopted solution for outdoor robot localization.

Point features are widely used in Visual-Odometry (VO) \cite{a3} and Visual-inertial-Odometry (VIO) systems \cite{c3,c4,c5} where descriptors or optical flow are employed to track point features in 2D images for camera pose estimation. However, in low-texture and dynamic environments, such as congested urban roads or highways, obtaining sufficient reliable visual point features is challenging. To address this limitation, incorporating additional structural features, such as lines, provides more constraints for state estimation. Similar to point features, line features require detection, tracking, and triangulation, all of which demand additional computational resources and pose challenges to real-time performance \cite{c8}. Furthermore, line triangulation is particularly susceptible to certain degenerate motions, especially in 2D motion, further complicating the use of line features in ground robots \cite{c9}.

Another approach to enhancing localization accuracy in outdoor environments is the removal of dynamic point features. A common method is to employ object detection algorithms to generate image masks that identify dynamic objects \cite{c21}. Point features within masked regions are excluded from the visual update process. However, this technique is computationally demanding and susceptible to missed detections. As an alternative, Motion Consistency Check (MCC) assesses the temporal consistency of point features over a short time window. Points that fail MCC are classified as dynamic and subsequently excluded from visual updates, improving the accuracy of estimation.

\begin{figure}
    \centering
    \includegraphics[width=1.0\linewidth]{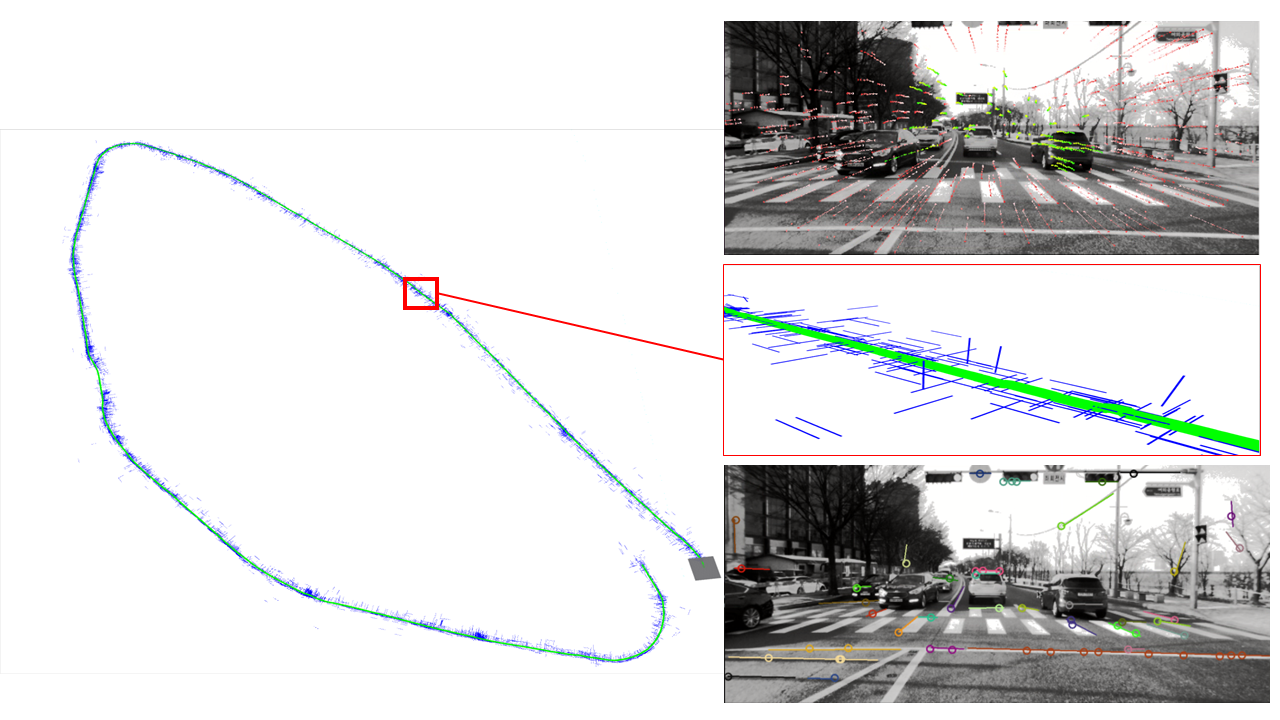}
    \caption{PL-VIWO in KAIST Urban32. $\textbf{Left}$: Top-down view of the trajectory with line mapping results. $\textbf{Top-Right}$: Point tracking results, where red dots connected by lines represent observations of point features within the sliding window, and green indicates those failed in triangulation or MCC. $\textbf{Middle-Right}$: Zoomed-in trajectory (green) with lines (blue). $\textbf{Bottom-Right}$: Point-line pairing results, where circles with the same color line indicate point features that lie on the corresponding line feature.}
    \label{fig:enter-label}
\end{figure}
To solve the aforementioned challenges, we propose a novel monocular visual-wheel-inertial odometry (VIWO) system tailored for ground robots navigating in outdoor scenarios. To enhance visual constraints in texture-less and highly dynamic scenarios, we integrate line features with an efficient processing method. The proposed pipeline exploits geometric point-line relationships in 2D images for faster line matching while avoiding degenerate motion issues in triangulation. This approach not only accelerates line feature processing but also ensures robust line triangulation. Furthermore, during point feature selection, we employ the MCC to filter out potential dynamic points. The contributions of this work can be summarized as follows:
\begin{itemize}

\item 
A filter-based tightly coupled monocular visual-inertial-wheel odometry system for accurate localization in complex outdoor environments is proposed. Our novel pipeline leverages point-line relationships to enhance real-time performance and robustness.


\item 
MCC is incorporated into our MSCKF-based visual update framework to maximize the utilization of valid point features in dynamic scenes.

\item Comprehensive experiments demonstrate that our system achieves superior accuracy, robustness and efficiency compared to existing state-of-the-art methods. 

\end{itemize}

\section{Related Works}
In this section,  related works are presented and categorised into: Visual-Inertial-Wheel Odometry, Line-aided VIO, and VIO in dynamic environments.
\subsection{Visual-Inertial-Wheel-Odometry}
Visual-inertial sensors are among the most widely used sensor combinations for mobile robot state estimation, valued for their complementary characteristics and cost-effectiveness. Numerous representative works have been proposed in recent years, which can generally be categorized into two approaches: filter-based and optimization-based. Prominent filter-based VIO systems, such as S-MSCKF \cite{c11}, and OpenVINS \cite{c3}, employ Kalman filters to update system states. In contrast, optimization-based methods, including VINS-Mono \cite{c4}, and ORB-SLAM3 \cite{c5}, formulate state estimation as an optimization problem and solve for the optimal system state through iterative refinement.

In VIO system for ground mobile robots, certain movements introduce additional unobservable states, leading to degraded estimation accuracy \cite{c2}. Integrating with wheel odometer addresses this issue by making the scale observable and significantly improving estimation accuracy. Lee et al. \cite{c13} tightly coupled the wheel with the VIO system based on the MSCKF framework, incorporating online calibration. Similarly, \cite{c14} utilized an optimization-based approach and introduced a robust initialization method leveraging wheel measurements. Other studies \cite{c15,c16,c27} have further explored planar motion constraints to improve state estimation of ground robots moving in 2D environments.

\subsection{Line-aided VIO}
Line features provide additional stable and robust geometric information, allowing the VO or VIO system to improve accuracy and robustness \cite{c9,c17,c18}. However, maintaining real-time performance with line features is often challenging. PL-VINS \cite{c26} addresses this by separating feature extraction and optimization into different threads, enabling real-time line-aided VIO. Based on this, \cite{c19} proposed PLF-VINS which use point-line coupled and parallel lines relation in the optimization process to enhance accuracy. Additionally, structured line features have demonstrated effectiveness in specific scenarios by leveraging horizontal and vertical lines to impose additional geometric constraints \cite{c30,c31}. These methods handle points and lines independently, disregarding the relationship between them. AirVo \cite{c20} introduced a line feature pipeline that utilizes learning-based point feature tracking for line tracking and triangulation, improving performance in illumination-challenging environments. Inspired by their work, we propose a novel method that integrates point and line features with their geometry relationship in 2D images, enabling efficient line tracking and robust triangulation in complex outdoor scenarios.

\subsection{VIO in Dynamic Enviroment}
In VIO systems, dynamic points often lead to inaccurate estimations.  Benefiting from deep learning technology, some works employ semantics segmentation algorithms to classify the potential dynamic points in 2D images \cite{c21,c22}. IMU integration which provides short-term 3D position estimation results can serve as a reference for MCC \cite{c23}. Similarly, Ground-Fusion \cite{c24} uses the wheel pre-integration results as the reference to filter the dynamic features. However, relying solely on a single sensor presents several challenges. For instance, IMU accelerometer noise accumulates significantly after double integration while wheel measurements are limited to 2D transformation information. Therefore, we utilize the historical poses in the sliding window as the reference for MCC to enhance system performance in dynamic scenes.

\section{Overview}
\begin{figure*}[t]
    \centering
    \includegraphics[width= 1.0\textwidth]{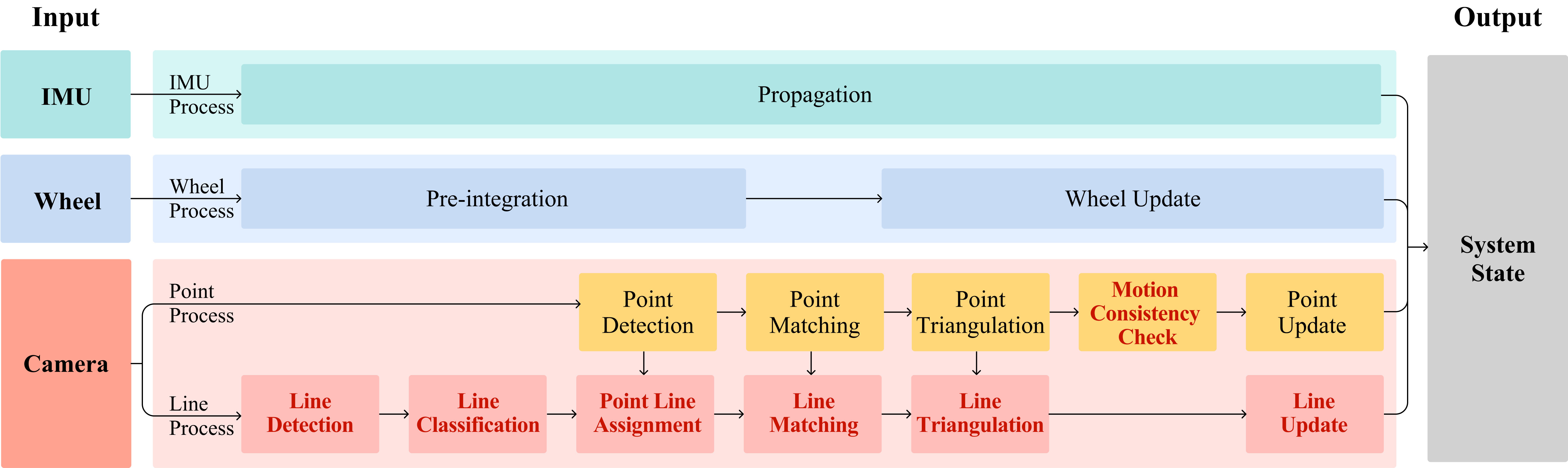}
    \caption{The framework of PL-VIWO system is divided into four components, each represented by different color-coded boxes. The boxes highlighted with bold red text indicate the key contributions of this work.} 
    \label{Framework}
\end{figure*}
For clarity of proposed method, this section first defines the system state, then briefly introduces the MSCKF algorithm, and finally presents the proposed system framework.

\subsection{System Dynamic Model}
Similar to the classical MSCKF-Based VINS system \cite{c3}, the state vector of the proposed system consists of current IMU state $\textbf{x}_{I_{k}}$ and  $n$ historical IMU clones $\textbf{x}_{H_{k}}$. 
\begin{align}
\textbf{x}_{I_{k}} &= (^{I_{k}}_{G}\textbf{R}, \ ^{G}\textbf{p}_{I_k}, \ ^{G}\textbf{v}_{I_k}, \ \textbf{b}_g, \ \textbf{b}_a) \\
\textbf{x}_{H_{k}} &= (^{I_{k-1}}_{G}\textbf{R}, \ ^{G}\textbf{p}_{I_{k-1}}, \ \dots, \ ^{I_{k-n}}_{G}\textbf{R}, \ ^{G}\textbf{p}_{I_{k-n}})
\end{align}

where $^{B}_{A}\textbf{R}$ is the rotation matrix from frame $A$ to $B$ and $ ^{A}\textbf{p}_{B}$ is position of $B$ in $A$. The terms $\textbf{b}_g$ and $\textbf{b}_a$ denote the gyroscope and accelerometer bias, respectively. In this work, we define the frames as follows: $G$ is the global frame, $I$ is the IMU frame, and $C$ corresponds to the camera frame.

For the Inertial-aided Navigation System (INS), IMU measurements $\textbf{w}_{m_{k}}$ and $\textbf{a}_{m_{k}}$ at timestamp $t_{k}$ are used to propagate the state from $t_k$ to $t_{k+1}$. 
\begin{equation}
\textbf{x}_{I_{k+1}} = \textbf{f}(\textbf{x}_{I_{k}}, \ \textbf{w}_{m_{k}}, \ \textbf{a}_{m_{k}})
\end{equation}

\subsection{MSCKF based Visual Update}
For feature measurements, the error state model can be generically represented as :
\begin{equation}
\tilde{\textbf{z}} = \textbf{z} - \textbf{h}(\hat{\textbf{x}}_{I_{k}}, {}^{G}\hat{\textbf{x}}_{f})
\end{equation}
where  $\hat{\textbf{x}}_{I_{k}}$ is the estimated state and $^{G}\hat{\textbf{x}}_{f}$ is the estimated feature position vector in the global frame. Then this error modal can be linearized as:
\begin{equation}
    \tilde{\textbf{z}} = \textbf{H}_{x} \hat{\textbf{x}}_{I_{k}} + \textbf{H}_{f} {}^{G}\hat{\textbf{x}}_{f} + \textbf{n}_{z}
\end{equation}
where $\textbf{H}_x$ and $\textbf{H}_f$ are the Jacobians with respect to system state $\textbf{x}_{I_{k}}$ and feature $^{G}\textbf{x}_{f}$, respectively. And $\textbf{n}_{z}$ is the measurement noise. By decomposing $\textbf{H}_f$, a new measurement modal independent of feature error can be derived :
\begin{equation}
\begin{aligned}
    \tilde{\textbf{z}}' &= \textbf{Q}_{n}^{T} \tilde{\textbf{z}} 
    \\ &= \textbf{Q}_{n}^{T} \textbf{H}_x \hat{\textbf{x}} + \textbf{Q}_{n}^{T} \textbf{H}_{f} {}^{G}\hat{\textbf{x}}_{f} + \textbf{Q}_{n}^{T} \textbf{n}_{f}
    \\ &= \textbf{H}'_x \hat{\textbf{x}}_{I_{k}} + \textbf{n}'_{z}
\end{aligned}
\end{equation}
where $\textbf{Q}_{n}^{T}$ is the null space of $\textbf{H}_{f}$ and $\textbf{Q}_{n}^{T} \textbf{H}_{f} = \textbf{0}$. Then the standard EKF update can be performed. 

\subsection{System Overview}



The proposed framework is illustrated in Fig. \ref{Framework}. Our PL-VIWO system consists of four main components: IMU, Wheel, Point, and Line Process. Built upon MINS \cite{c1}, our system retains the same IMU and Wheel Process. For the point update, we integrate the MCC module to maximize the inclusion of reliable visual point features in dynamic environments. Additionally, line features are incorporated to reinforce visual constraints with a novel pipeline which exploits the relationship between point and line features for efficient line tracking. To address the challenge of degenerate motions in ground robots, which often prevent line triangulation, we propose a robust line triangulation method that leverages the point-line relationship and point triangulation results. For the line update, MSCKF is employed in the same manner as for the point update.
\begin{figure*}[t]
    \centering
    \subfloat[Line detection]{\includegraphics[width=0.45\textwidth]{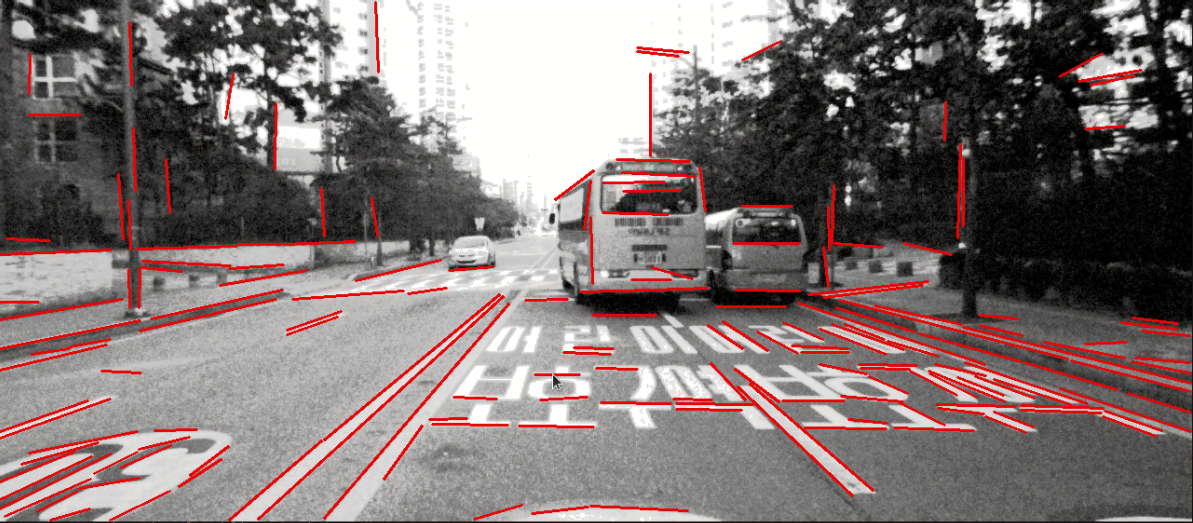}}
    \subfloat[Line classification]{\includegraphics[width=0.45\textwidth]{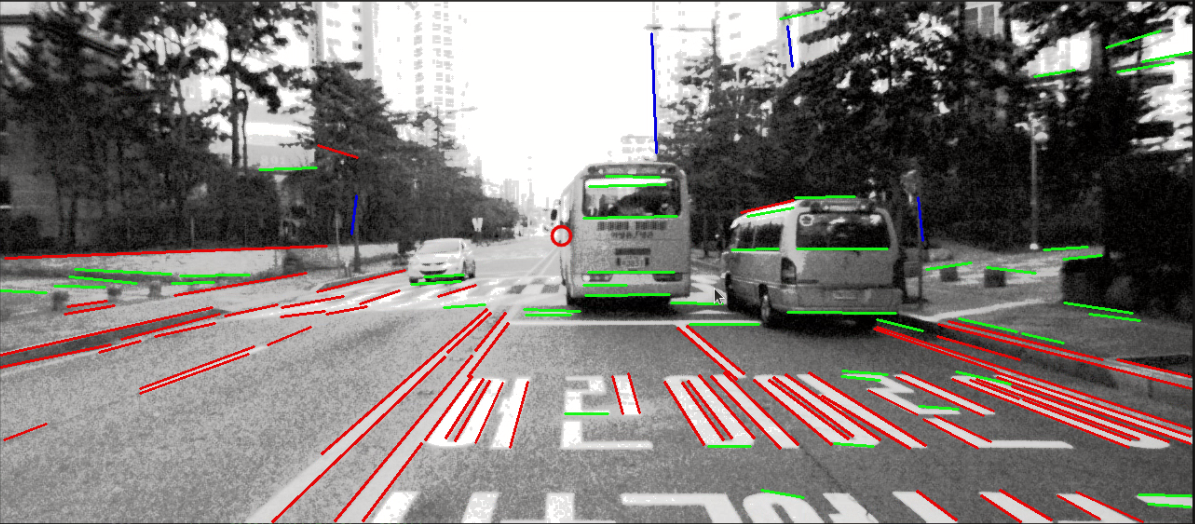}} \\
    \subfloat[Point line assignment]{\includegraphics[width=0.45\textwidth]{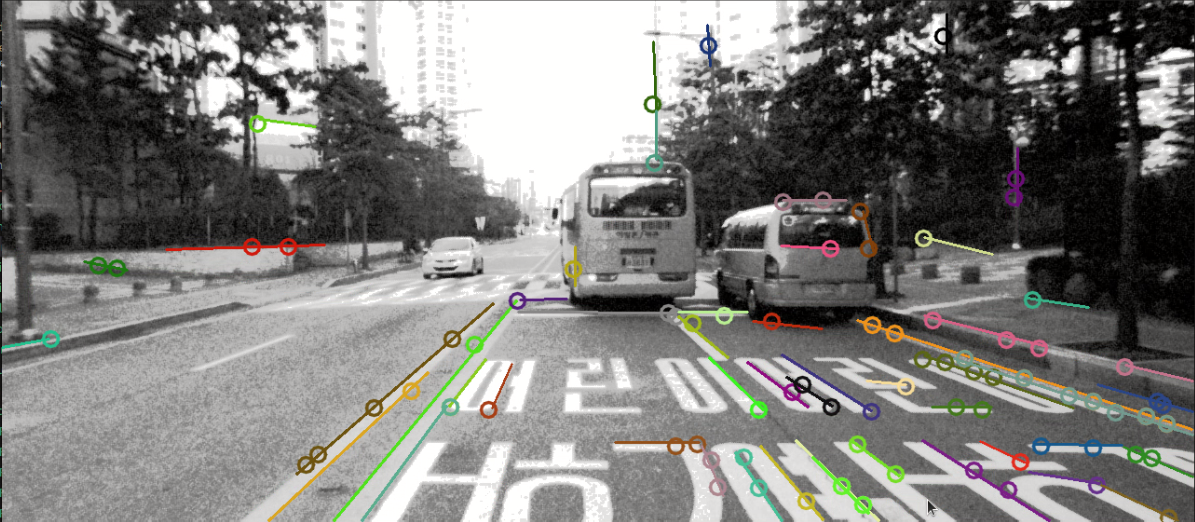}}
    \subfloat[Line matching]{\includegraphics[width=0.45\textwidth]{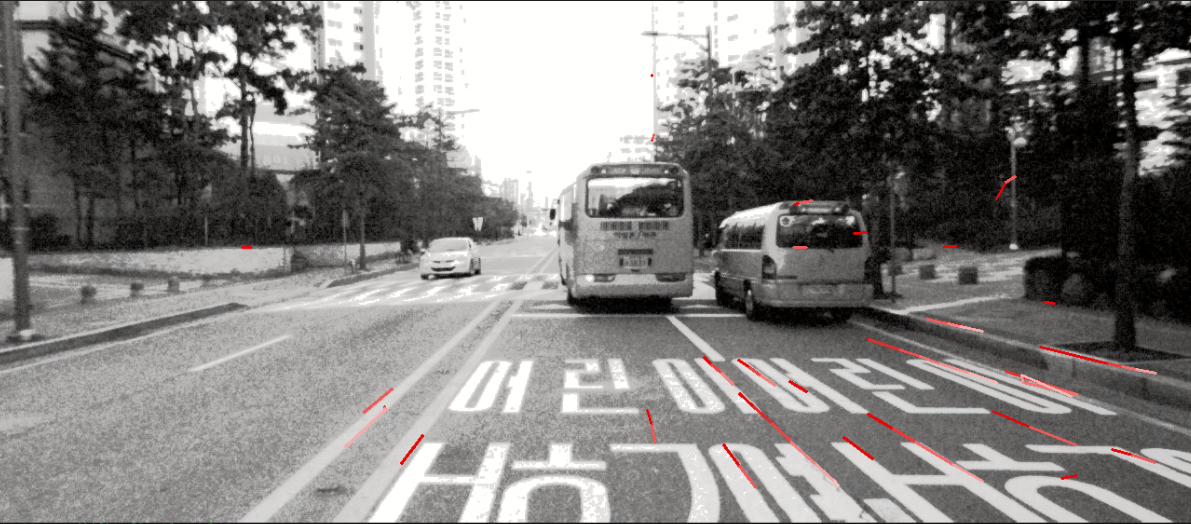}}
    \caption{2D Line process results in KAIST Urban 28. (a) The red line segments represent the detection result. 
    (b) The colored lines represent the classification results: red, green, and blue indicate lines parallel to the x-, y-, and z-axes of the IMU frame, respectively. Non-parallel lines are omitted from the diagram. The red circle around the centre of the image represents the vanishing point corresponding to the x-axis while y and z lie outside the image.
    (c) Circles lying on a line with the same color represent point features assigned with a line.
    (d) The colored lines in the figure are formed by connecting the midpoints of all observed line segments within the sliding window. The redder line means more recent.}
    \label{fig:2D_Line}
\end{figure*}

\section{Methodology}
Here, we describe the key components of the proposed approach, with a focus on the contributions.
\subsection{Motion Consistency Check} 
Outdoor ground robot localization often suffers from estimation inaccuracies due to dynamic objects. A common approach in MSCKF-based VIO to address this issue is to discard point features with large reprojection errors or those failing the Chi-square test during updates. However, in highly dynamic environments, this strategy may result in an insufficient number of valid point features for visual updates, even if numerous point features are extracted, which ultimately degrades localization accuracy.

In our system, we introduce MCC during the point feature selection to ensure a sufficient number of points are used for updates. This approach effectively increases the number of reliable point features for visual updates. Specifically, for a point feature $^{G}\textbf{p}_{f}$ with $n$ measurements, the average projection error $\textbf{r}$ is defined as:
\begin{equation}
    \textbf{r} = \frac{1}{n} \sum \limits_{i=1}^{n}\left\| \textbf{z}_{i} - \boldsymbol{\pi} ({}_{G}^{C_{i}}\textbf{R} {} (^{G}\hat{\textbf{p}}_{f} - {}^{G}\textbf{p}_{C_{i}})) \right\| 
\end{equation}
where $\textbf{z}_{i}$ is the $i_{th}$ image measurement and $^{G}\hat{\textbf{p}}_{f}$ is the 3D position of point feature derived from triangulation. The function $\boldsymbol{\pi}$ is the camera projection function. For feature points whose error $\textbf{r}$ are greater than the threshold, they are considered dynamic points and excluded from the update.

\subsection{2D Line Process} 
The pipeline for 2D line process consist of Line Detection, Line Classification, Point-Line Assignment and Line Matching. 
\subsubsection{Line Detection}
Line Segment Detector (LSD) \cite{b24} is a classical 2D line detection algorithm that identifies the two endpoints of all line segments in an image. Since the line detection speed is influenced by the image size, the image is downsampled to 0.25 times to enhance efficiency in our system. Additionally, as detected line segments vary in length, longer segments tend to be more accurate and easier to track. Therefore, we discard the short line segments and the detection result is shown in Fig. \ref{fig:2D_Line} (a).

\subsubsection{Line Classification}
\label{subsubsection:2}
The detected 2D line segments correspond to straight lines in 3D space with various orientations. However, lines aligned with the motion direction suffer from degenerate motion, preventing reliable triangulation via plane intersections from consecutive frames \cite{c9}. Therefore, we utilize vanishing points to classify 2D line segments based on their parallelism to different IMU frame axes. During triangulation, the classified lines are assumed to align with their respective coordinate axes.

The vanishing point is the point where parallel lines in 3D space appear to converge in a 2D perspective projection which is widely used to identify parallel relationships in 3D space from 2D images \cite{c30,c31}. The vanishing points $\textbf{vp}_{x}$ along x-axis, can be formulated as:
\begin{equation}
    \textbf{vp}_{x} = \boldsymbol{\pi}({}_{I}^{C}\textbf{R} \textbf{u}_{x})
\end{equation}
where $_{I}^{C}\textbf{R}$ represents the extrinsic rotation matrix and $\textbf{u}_{x}$ is the unit vector along x-axis. The same calculation method is applied to obtain the vanishing point for the y and z-axis. By connecting the midpoints of the detected line segments to different vanishing points, new straight lines are formed. The alignment quality is then evaluated by computing the angle error $e_{angle}$ and distance error $e_{dist}$ between these generated lines and the originally detected segments as follows:
\begin{equation}
    {e}_{angle} = \text{atan}(\frac{\textbf{p}_{s}(1) - \textbf{p}_{e}(1)}{\textbf{p}_{s}(0) - \textbf{p}_{e}(0)}) - 
    \text{atan}(\frac{\textbf{p}_{m}(1) - \textbf{vp}(1)}{\textbf{p}_{m}(0) - \textbf{vp}(0)})
\end{equation}{}
\begin{equation}
    e_{dist} = \frac{\textbf{n}^\top \begin{bmatrix} \textbf{p}_{s}(0) & \textbf{p}_{s}(1) & 1 \end{bmatrix} + \textbf{n}^{\top} \begin{bmatrix} \textbf{p}_{e}(0) & \textbf{p}_{e}(1) & 1 \end{bmatrix}} 
    {2 \sqrt{{\textbf{n} (0)}^{2} + {\textbf{n} (1)}^{2}}}
\end{equation}
where $\textbf{p}_{s}$, $\textbf{p}_{e}$, $\textbf{p}_{m}$ are the startpoint, endpoint and midpoint, respectively. And $\textbf{n}$ is the norm vector formed by vanishing point and midpoint:
\begin{equation}
    \textbf{n} = \begin{bmatrix} \textbf{p}_{m}(0) & \textbf{p}_{m}(1) & 1 \end{bmatrix}^{\top}  \times
    {\begin{bmatrix} \textbf{vp}(0) & \textbf{vp}(1) & 1 \end{bmatrix}}^{\top}
\end{equation}
A line segment and a vanishing point are considered collinear only if both errors are below the threshold. The classification result is shown in Fig. \ref{fig:2D_Line}(b).
\begin{figure*}[t]
    \centering
    \subfloat[]{\includegraphics[width=0.33\textwidth]{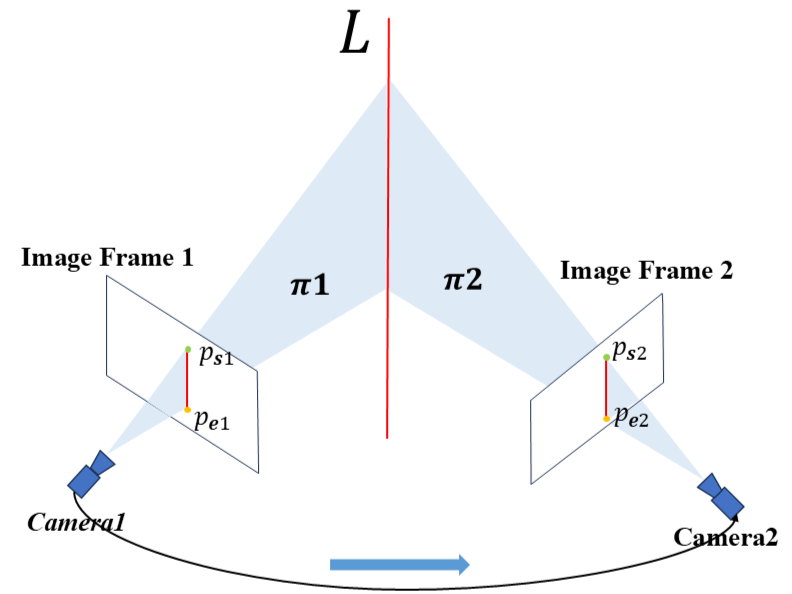}}
    \subfloat[]{\includegraphics[width=0.33\textwidth]{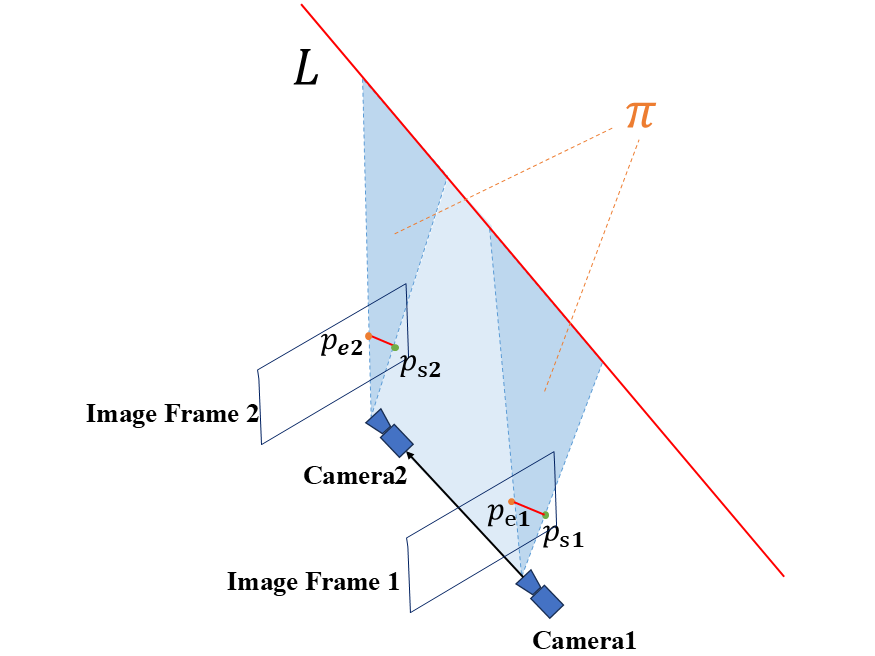}}
    \subfloat[]{\includegraphics[width=0.33\textwidth]{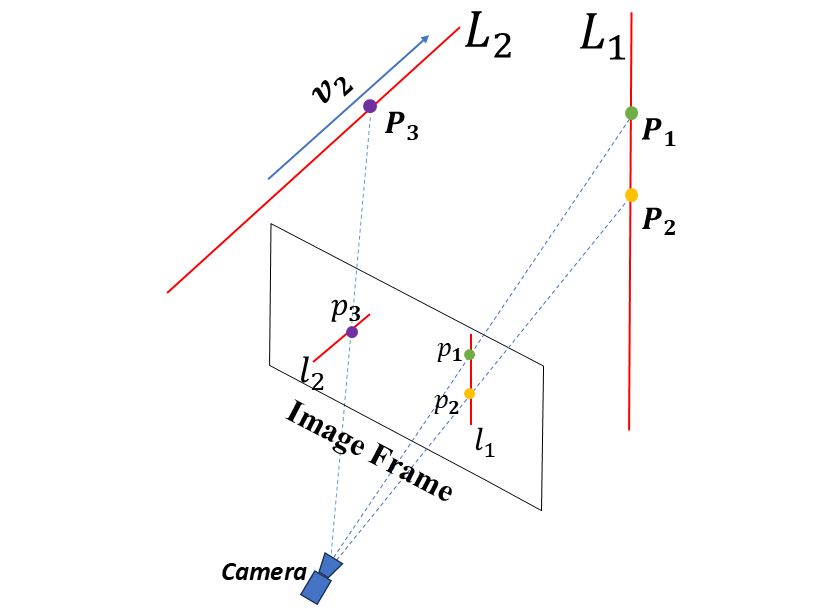}}
    \caption{ Line triangulation
    (a) Triangulation from planes. 
    (b) Degenerate motion for line feature triangulation. Cameras stay in the same plane $\pi$.
    (c) Triangulation from points and direction.
    }
    \label{fig:3D_line}
\end{figure*}

\subsubsection{Point-Line Assignment}
\label{subsubsection:3}
By analyzing the distance between points and lines in a 2D image, the correspondence between them can be established directly. Since the points tracking results have already been obtained during the point feature update, the tracking results of line features can be determined based on the associated point tracking results. This approach eliminates the need for line feature descriptors, significantly accelerating line matching. 

For point $\textbf{p}$:${\begin{bmatrix} u_p & v_p \end{bmatrix}}^{\top}$ and line $\textbf{l}$:$\begin{bmatrix} u_s & v_s & u_e & v_e    
\end{bmatrix}^{\top}$ on 2D image, the distance between them can be computed by:
\[
\mathit{d} =
\begin{cases} 
\sqrt{(u_p - u_s)^2 + (v_p - v_s)^2}, & \text{if } \mathit{cross} \leq 0 \\[8pt]
\sqrt{(u_p - u_e)^2 + (v_p - u_e)^2}, & \text{if } \mathit{cross} > \mathit{len}^2 \\[8pt]
\frac{|(v_e - v_s)u_p + (u_s - u_e)v_p + (u_e v_s - u_s v_e)|}{\sqrt{(v_e - v_s)^2 + (u_s - u_e)^2}}, & \text{otherwise}
\end{cases}
\]
where $\mathit{cross}$ is the projection of the point on the line segment and $\mathit{len}$ is the length of line segment.
\begin{equation}
\begin{aligned}
    \mathit{len} &= \sqrt{{(u_s - u_e)^{2} + (v_s - v_e)^{2}}}  \\
    \mathit{cross} &= (u_e - u_s)(u_p - u_s) + (v_e - v_s)(v_p - v_s)
\end{aligned}
\end{equation}
For $\mathit{cross} < 0$, the point project on the left side of the line startpoint $\textbf{p}_{s}$. Conversely, its projection lies on the right side of endpoint $\textbf{p}_{e}$ when $cross> d^2$. A 2D point is considered to be on a line only if its projection falls between the two endpoints and the distance to the line is less than $3$ pixels. The point line assignment result is shown in Fig. \ref{fig:2D_Line}(c). 

\subsubsection{Line Matching}
The line features can be tracked by using point matching and point line assignment results. AirVO \cite{c20} scores matches based on the number of points which is unsuitable for outdoor complex scenes. A possible reason is that line features in such environments are scattered and numerous, making it unlikely for multiple lines to share the same points. Therefore, we adopt two simpler rules:
\begin{itemize}
  \item Two line segments in adjacent frames are considered the same 3D line if they share at least two point features.
  \item If only one point is assigned, the differences in position and direction between the two line segments are calculated. If both differences are below the threshold, they are considered to be the same. (We assume a line maintains the same direction and undergoes limited positional changes between adjacent frames).
\end{itemize}
Fig. \ref{fig:2D_Line}(d) shows the line matching results.

\subsection{Line Triangulation}
This paper uses the Plücker coordinate to represent a 3D spatial line $\textbf{L}$, as it facilitates spatial transformations. It consists of a normal vector $\textbf{n}$ and a direction vector $\textbf{v}$.
\begin{equation}
    \textbf{L} = \left[\begin{array}{c c}
    \textbf{n} & \textbf{v}
\end{array}\right] ^\top
\end{equation}

Classical line triangulation methods \cite{c9} project 2D line segments into 3D as planes. The intersection of planes forms the 3D line, as illustrated in Fig. \ref{fig:3D_line}(a). However, this triangulation method suffers from degenerate motion. When the camera moves parallel to the line direction, the two formed planes coincide making it impossible to recover the spatial line, like in Fig. \ref{fig:3D_line} (b). This issue is particularly common in 2D urban navigation for ground robots, where line features such as lane markings and curbs cannot be triangulated when the robot moves in a straight line. Consequently, even if most line features are successfully tracked, they cannot be used for visual updates without accurate triangulation.

To address this, we introduce two novel line triangulation methods to handle the cases when classical triangulation fails. Since points and lines are associated in \ref{subsubsection:3}, and the direction of some lines has been determined in \ref{subsubsection:2}, this additional information is leveraged in our methods. Firstly, if more than two triangulated points are assigned to a line $\textbf{L}$, then choose the closest two (ie. $\textbf{p}_{1}$ and $\textbf{p}_{2}$) and a 3D line can be obtained, as illustrated in Fig. \ref{fig:3D_line}(c) right part.
\begin{equation}
    \textbf{L} = \left[\begin{array}{c c}
    \textbf{n} \\
    \textbf{v}
    \end{array}\right] = 
    \left[\begin{array}{c c}
    \textbf{p}_{1} \times \textbf{p}_{2} \\
    \frac{\textbf{p}_{1} - \textbf{p}_{2}} {\left | {\textbf{p}_{1} - \textbf{p}_{2}} \right |}
    \end{array}\right] ^\top 
\end{equation}
However, triangulating points in outdoor dynamic scenes is challenging, there are cases where only one point $\textbf{p}_{3}$ has been triangulated. In such cases, the second triangulation method will be used. If the line direction is parallel to x-axis of the IMU coordinate, the direction vector can be directly obtained from the rotation matrix ${}_{G}^{I}\textbf{R}$, as shown in Fig. \ref{fig:3D_line}(c) left:
\begin{equation}
    \textbf{v}_{x} = {}_{G}^{I}\textbf{R} \textbf{u}_{x}
\end{equation}
Where $\textbf{u}_{x}$ is the unit vector along x-axis. Lines parallel to y and z-axis can be determined in the same way. Then the norm vector $\textbf{n}$ can be derived by using point $\textbf{p}_{3}$:
\begin{equation}
    \textbf{n} = \textbf{p}_{3} \times \textbf{v}
\end{equation}

\subsection{Line Measurement Model}

The line measurement model describes the distance between the observed two endpoints $\textbf{p}_{s}$ and $\textbf{p}_{e}$, to the projected line $\mathbf{l}$ in the 2D image coordinate:
\begin{equation}
\mathbf{z_l} = \left[\begin{array}{cc}
    \mathbf{d}_{s} & \mathbf{d}_{e} 
\end{array}\right]^{\top} =
\left[\begin{array}{c c}
\frac{\mathbf{p}_s^\top \mathbf{l}}{\sqrt{l_1^2 + l_2^2}} & \frac{\mathbf{p}_e^\top \mathbf{l}}{\sqrt{l_1^2 + l_2^2}}
\end{array}\right] ^\top
\label{eq:line_measurement}
\end{equation}
where $\mathbf{p_s} = \begin{bmatrix} u_s& v_s&1 \end{bmatrix}^\top$ and $\mathbf{p_e} = \begin{bmatrix} u_e& v_e&1 \end{bmatrix}^\top$. The projected line $\mathbf{l}$ can be derived by the camera pose and  Plücker coordinates of 3D line in the global frame $^{G}\textbf{L}$:

\begin{equation}
\begin{aligned}
\mathbf{l} &= \left[\begin{array}{c c}
\mathbf{K} & \mathbf{0}_3
\end{array}\right]{^{G}\mathbf{L}}
= \left[\begin{array}{c c}
\mathbf{K} & \mathbf{0}_3
\end{array}\right] \left[\begin{array}{c c} ^{G}{\mathbf{n}} & ^{G}{\mathbf{v}} \end{array}\right]^{\top}\\
&= \left[\begin{array}{c c} \mathbf{K} & \mathbf{0}_3 \end{array}\right]
\left[\begin{array}{c c} _{G}^{C}{\mathbf{R}} & [^{C}{\mathbf{p}}_{G}]_\times {_{G}^{C}{\mathbf{R}}}\\ \mathbf{0}_3 & _{G}^{C}{\mathbf{R}} \end{array}\right]
\left[\begin{array}{c c} ^{G}{\mathbf{n}} & ^{G}{\mathbf{v}} \end{array}\right]^{\top}\\
\end{aligned}
\end{equation}
where $\mathbf{K}$ is camera intrinsic matrix for line projection and $[\textbf{x}]_{\times}$ represent the skew-symmetric matrix of $\textbf{x}$. The Jacobian matrices of the measurement model with respect to the IMU pose and line features follow a similar formulation as presented in the Appendix of \cite{c9}.
\section{EXPERIMENTS}
We evaluated the PL-VIWO using the publicly available KAIST Complex Urban Dataset \cite{c25}. To assess the computational efficiency and localization accuracy, we compared the proposed PL-VIWO against state-of-the-art monocular VIO and VIWO methods. The benchmark includes three optimization-based algorithmes: VINS-Mono\cite{c4}, PL-VINS\cite{c26}, and VIW-Fusion\cite{c27}, as well as filter-based: MINS\cite{c1}. Since PL-VIWO is built on MINS with MCC and line features, MINS serves as a key benchmark to highlight the effectiveness of our methods. PL-VINS is a VIO system built upon VINS-Mono integrated line features, while VIW-Fusion is a VIWO system that incorporates the wheel information into VINS-Mono. All experiments run on an Intel Core i7 CPU, 32 GB of RAM, and an NVIDIA A1000 GPU (6 GB VRAM) with Ubuntu 20.04.

Note that we initially used the same IMU parameters for both types which resulted in suboptimal performance for the optimization-based algorithms. To ensure fair and reliable evaluation, we tuned IMU noise in optimization-based methods for their best performance. Up to 150 points were used for visual updates. Additionally, we disabled the online calibration of both internal and external parameters, as well as the loop closure in all algorithms for fairness. 

\subsection{Localization Accuracy}
\begin{figure}[!t]
\centering
\includegraphics[width=0.45\textwidth]{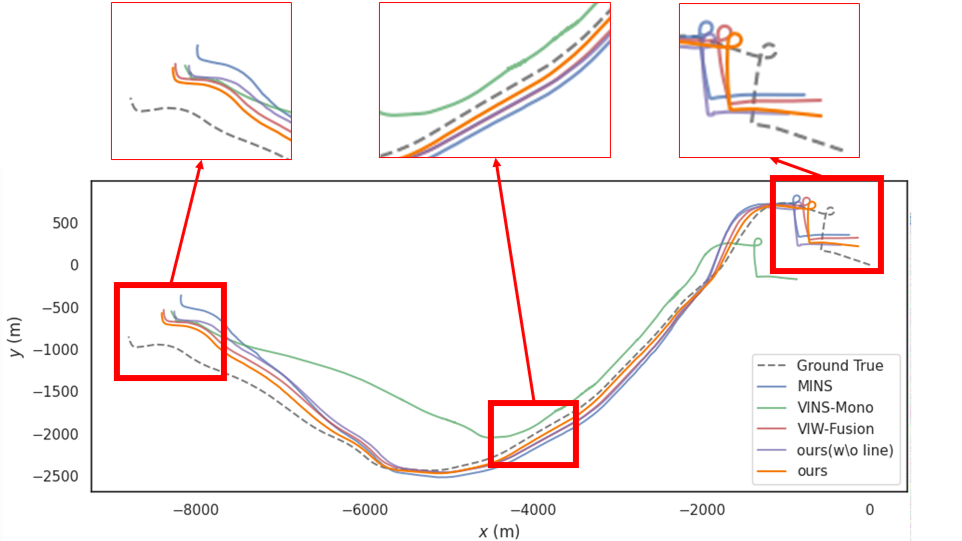}
\caption{Trajectories of the KAIST Urban 31(PL-VINS failed).} 
\label{Trajectories}
\end{figure}

\begin{table*}
\centering
\renewcommand\arraystretch{1.2} 
	\caption{RMSE ATE: POSITION($m$)$\And$ORIENTATION($^{\circ}$) FOR ALGORITHMS IN HIGHWAY SEQUENCES.}
	\label{table1}
	\setlength{\tabcolsep}{0.5pt} 
	\begin{tabular}{|m{3.6cm}<{\centering}|m{1.2cm}<{\centering}|m{1.2cm}<{\centering}|m{1.2cm}<{\centering}|m{1.2cm}<{\centering}|m{1.2cm}<{\centering}|m{1.2cm}<{\centering}|m{1.2cm}<{\centering}|m{1.2cm}<{\centering}|m{1.2cm}<{\centering}|m{1.2cm}<{\centering}|m{1.2cm}<{\centering}|}
		\hline 
		Algorithms & urban18 (3.9km) & urban19 (3.0km) & urban20 (3.2km) & urban21 (3.7km) & urban22 (3.4km) & urban23 (3.4km) & urban24 (4.2km) & urban25 (2.5km) &urban35 (3.2km) & urban36 (9.0km) & urban37 (11.8km) \\  \hline \hline
		 \rowcolor{gray!30} {VINS-mono} & {FAIL} & {FAIL} & {FAIL} & {FAIL} & {FAIL} & {FAIL} & {FAIL} & {FAIL} & {FAIL} & {FAIL} & {FAIL} \\
             \rowcolor{gray!30} & {} & {} & {} & {} & {} & {} & {} & {} &  & {} & {} \\ \hline
		\rowcolor{gray!30} {PL-VINS} & {FAIL} & {FAIL} & {FAIL} & {FAIL} & {FAIL} & 208.26 & {FAIL} & {FAIL} & 256.70 & {FAIL} & {FAIL} \\
             \rowcolor{gray!30}   & &  &  &  &  & 7.10 &  &  & 159.09 &  &  \\ \hline  
		\rowcolor{gray!30} {VIW-Fusion} & \textbf{43.60} & 37.57 & \underline{38.65} & 52.86 & 53.05 & 42.28 & 55.88 & \textbf{29.46} & \underline{36.77} & 116.98 & \underline{174.87} \\
                 \rowcolor{gray!30}    & 4.27 & 6.80 & 155.40 & 160.25 & 156.0 & 12.00 & 55.12 & \textbf{9.01} & \underline{10.35} & 20.02 & 158.55 \\ \hline
		{MINS(I,C,W)} & 50.58 & 37.19 & 40.67 & \underline{46.33} & \textbf{46.12} & \underline{42.53} & \textbf{47.65} & 36.60 & 55.01 & 117.89 & 197.53 \\
           -mono & 2.94 & 6.02 & \underline{9.39} & 12.67 & 16.71 & 5.95 & 36.93 & 37.30 & 15.68 & 6.34 & 4.16 \\ \hline
           
            {PL-VIWO} & 52.50 & \underline{37.08} & 40.40 & 47.37 & \underline{48.55} & 43.00 & \underline{48.76} & 31.38 & 40.01 & \underline{116.63} & 178.50\\
            (w/o line) & \underline{2.86} & \underline{3.27} & 10.18 & \textbf{6.06} & \textbf{15.30} & \underline{2.01} & \underline{24.57} & 28.98 & {10.37} & \underline{4.85} & \underline{4.05}\\ \hline
             
            {PL-VIWO} & \underline{47.99} & \textbf{36.18} & \textbf{36.86} & \textbf{46.15} & 49.77 & \textbf{41.10} & 51.32 & \underline{31.01} & \textbf{36.09} & \textbf{105.46} &  \textbf{136.04}\\
             & \textbf{2.80} & \textbf{3.22} & \textbf{6.74} & \underline{6.43} & \underline{15.80} & \textbf{1.00} & \textbf{24.15} & \underline{16.80} & \textbf{8.51} & \textbf{4.40} &  \textbf{3.76}\\ \hline

        \end{tabular}
	\label{tab:Highway}
\end{table*}
\begin{table*}[h]
\centering
\renewcommand\arraystretch{1.2} 
	\caption{RMSE ATE: POSITION($m$)$\And$ORIENTATION($^{\circ}$) FOR ALGORITHMS IN URBAN SEQUENCES.}
	\label{table2}
	\setlength{\tabcolsep}{1pt} 
	\begin{tabular}{|m{3.6cm}<{\centering}|m{1.3cm}<{\centering}|m{1.3cm}<{\centering}|m{1.3cm}<{\centering}|m{1.3cm}<{\centering}|m{1.3cm}<{\centering}|m{1.3cm}<{\centering}|m{1.3cm}<{\centering}|m{1.3cm}<{\centering}|m{1.3cm}<{\centering}|m{1.3cm}<{\centering}|m{1.3cm}<{\centering}|}
		\hline 
		Algorithms & urban26 (4.0km) & urban27 (5.4km) & urban28 (11.5km) & urban29 (3.6km) & urban30 (6.0km) & urban31 (11.4km) & urban32 (7.1km) & urban33 (7.6km) &urban34 (7.8km) & urban38 (11.4km)\\      \hline \hline
		\rowcolor{gray!30} VINS-mono & 35.43 & 151.34 & 94.77 & FAIL & 110.40 & FAIL & \underline{74.24} & FAIL & FAIL & 170.39\\ 
             \rowcolor{gray!30} & 3.58 & 7.04 &  \textbf{3.58} &  & 7.35 &  & \textbf{5.90} &  &  & 8.40 \\ \hline 
             
		\rowcolor{gray!30} PL-VINS & FAIL & FAIL & FAIL & FAIL & FAIL & FAIL & 154.41 & 134.07 & 291.33 & 284.88 \\
               \rowcolor{gray!30} &  &   &  &  &   & & 8.83 & \underline{7.38} & 18.47 & 18.03  \\ \hline  
		
        \rowcolor{gray!30} VIW-Fusion & \textbf{23.86} & 124.01 & \textbf{32.28} & \textbf{42.26} & \textbf{40.72} & \underline{230.15} & 86.11 & 127.52 & 39.18 & \textbf{48.12} \\
                \rowcolor{gray!30} & \underline{3.31} & 32.35 & \underline{5.05} & \textbf{4.70} & \textbf{5.56} & 14.08 & \underline{7.03} & 16.63 & \textbf{3.62} & \underline{7.60} \\ \hline
		MINS(I,C,W) & 34.87 & 68.87 & 91.00 & 68.41 & 73.56 & 334.15 & 93.89 & \underline{85.47} & 40.64 & 104.04 \\
                -mono & 3.87 & \underline{6.79} & 11.79 & 9.80 & 10.37 & 16.83 & 9.45 & 8.20 & 6.00 & 10.78 \\ \hline
            {PL-VIWO} & 36.30 & \underline{55.24} & 73.54 & 66.10 & \underline{64.01} & 302.19 & 78.08 & \textbf{75.29} & \underline{37.99} & 96.16\\
            (w/o line) & 3.51 & 7.19 & 7.70 & 9.62 & \underline{5.54} & \underline{13.47} & 8.32 & 9.05 & 6.19 & 9.75 \\ \hline
            {PL-VIWO} & \underline{28.84} & \textbf{47.67} & \underline{49.19} & \underline{47.34} & 65.67 & \textbf{191.77} & \textbf{65.96} & 86.63 & \textbf{33.60} & \underline{73.31}\\
             & \textbf{2.27} & \textbf{3.03} & 7.00 & \underline{6.83} & 6.63 & \textbf{11.15} & 7.78 & \textbf{6.88} & \underline{5.54} & \textbf{6.98} \\ \hline
        \end{tabular}
	\label{tab:City}
\end{table*}
Due to the varying test scenarios, we divided the localization experiments into two groups: highway and city scenes. The localization accuracy is evaluated using the root mean squared error (RMSE) of the absolute pose (position and orientation), computed by EVO \cite{c28}. Optimization-based algorithms are distinguished with a grey background in the result tables. Meanwhile, position errors above 300m or test not completed the test are listed as $FAIL$. The best results are \textbf{highlighted} and the second best are \underline{underlined}.
\subsubsection{Highway} 
Table \ref{tab:Highway} shows the localization results for highway scenarios. Since motion on highways is mostly uniform and linear, the VINS system suffers from degenerate motion. Additionally, visual constraints are weak due to the lack of texture information in highway environments. These factors lead to poor performance for both VINS-Mono and PL-VINS. VINS coupled with wheel achieves higher accuracy due to the incorporation of wheel information. Compared with MINS, our system utilizes more valid point features during the visual update process, resulting in improved accuracy even without incorporating line features. Our system further improves and achieves the best accuracy in 8 out of 11 sequences after introducing line features.
\subsubsection{City}
The localization results in city scenarios are shown in Table \ref{tab:City}. On urban roads, due to frequent changes in vehicle speed and the abundance of texture, VINS-Mono can achieve relatively accurate results. However, its accuracy remains lower than the system coupled with wheel in most sequences. PL-VINS fails in most of the scenarios, possibly due to dynamic lines, such as lines on moving vehicles, being incorrectly triangulated and added to the optimization process. Compared with MINS, our system demonstrates significant improvements in both translation and rotation accuracy due to the introduction of more effective visual constraints. These improvements are further enhanced after incorporating line features. VIW-Fusion achieves the best results in several sequences due to its optimization-based framework, which enhances robustness at the cost of higher computation. Overall our approach is the best in 5 out of 10 sequences and second best in 4 out of 10 sequences. The trajectories of different algorithms in Ubran 31 are shown in Fig. \ref{Trajectories}. The trajectory from PL-VIWO with line is the closest to the ground truth among all methods, further validating the effectiveness of our method.
\begin{table}
    \centering
    \caption{Comparison of Average Feature Processing Time of Principal Component with PL-VINS in Urban38.}
    \begin{tabular}{lcccccc}
        \toprule
        & \makecell[c]{Point \\ Extraction} 
        & \makecell[c]{Line \\ Extraction} 
        & \makecell[c]{Point \\ Matching}
        & \makecell[c]{Line \\ Matching} 
        & \makecell[c]{Total \\ Time}  \\
        \midrule
        PL-VINS & 10 ms & 23 ms & 3.0 ms & 6 ms & 42 ms  \\
        PL-VIWO  & 5 ms & 7 ms & 2.5 ms & 0.5 ms & 15 ms \\
        \bottomrule
    \end{tabular}
    \label{tab:feature-time}
\end{table}

\subsection{Computation Efficiency Analysis}
\begin{figure}[!t]
\centering
\includegraphics[width=0.5\textwidth]{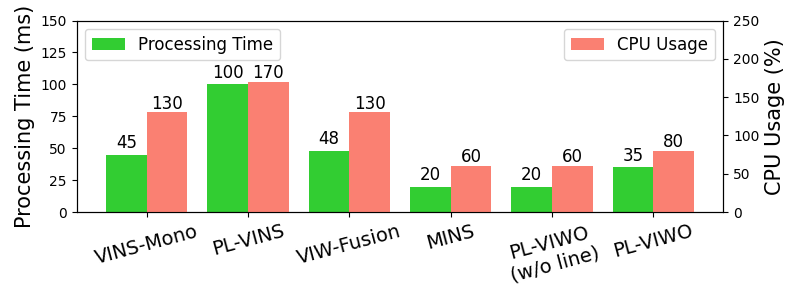}
\caption{Bar chat of average process time between frames and CPU usage across different algorithms.} 
\label{BarChat}
\end{figure}
To assess the computational efficiency of the proposed method, we analyze the following two modules.
\subsubsection{Feature Process Time}
First, we record and compare the processing time of 2D image feature process in our system with PL-VINS, as shown in Table \ref{tab:feature-time}. Our system achieves faster performance in point extraction and tracking by employing a grid-based extraction method, which prevents redundant feature extraction in certain grid regions. For line extraction, we downscale the image, reducing the processing time to 7ms. Additionally, our line-matching approach avoids descriptor computation, achieving a processing time of just 0.5ms, whereas PL-VINS takes 6ms due to descriptor calculation. Overall, our proposed pipeline demonstrates greater efficiency in 2D image processing.
\subsubsection{Total Process Time $\&$ CPU Usage}
Additionally, the processing time and CPU usage of each algorithm are recorded in Fig. \ref{BarChat}. For filtering-based algorithms, the processing time includes adjacent frame image processing and state updates. In contrast, in optimization-based algorithms, the processing time between adjacent keyframes consists of image processing and optimization. Filter-based algorithm frameworks require less processing time compared to optimization-based. Introducing line features and wheel data will add additional processing time. Regarding CPU usage, optimization-based methods employ additional threads for optimization. Consequently, their CPU usage is significantly higher than that of filtering-based algorithms, which run on a single thread. This makes our algorithm more suitable for deployment on resource-constrained edge devices.
\subsection{Line Triangulation Results}
\begin{figure}[t]
    \centering
    \subfloat[]{\includegraphics[width=0.25\textwidth]{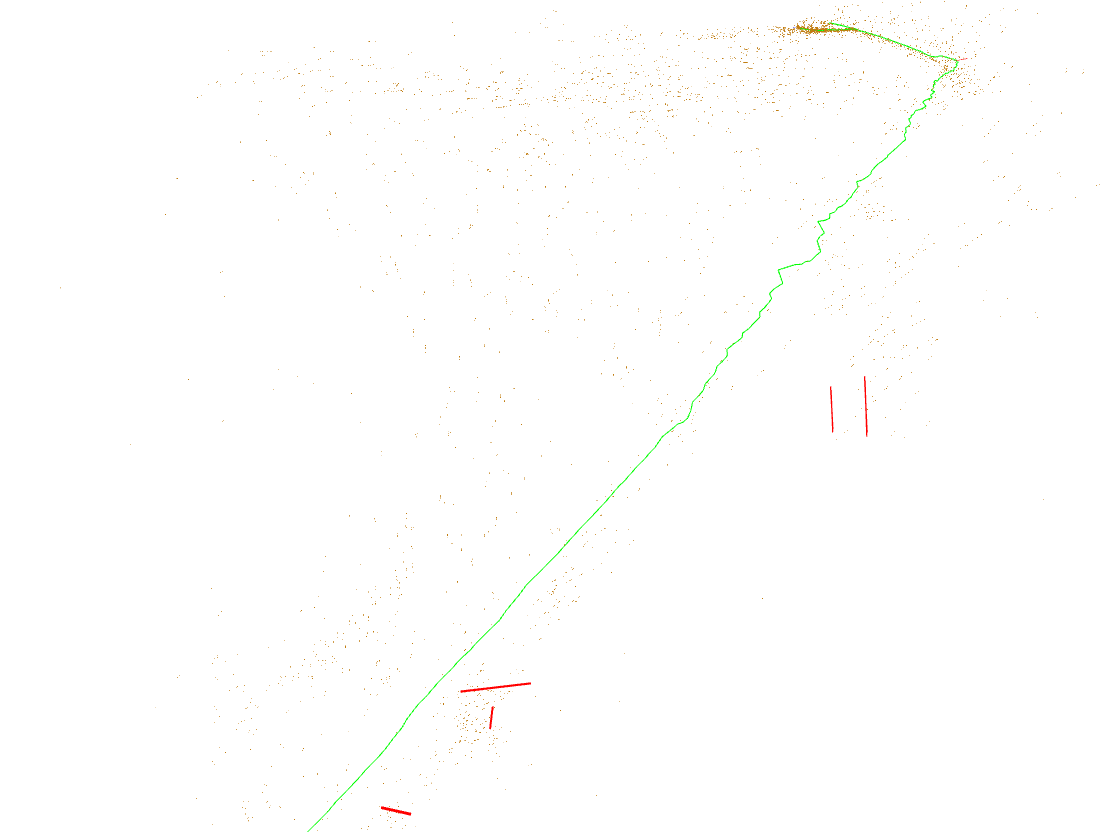}}
    \subfloat[]{\includegraphics[width=0.25\textwidth]{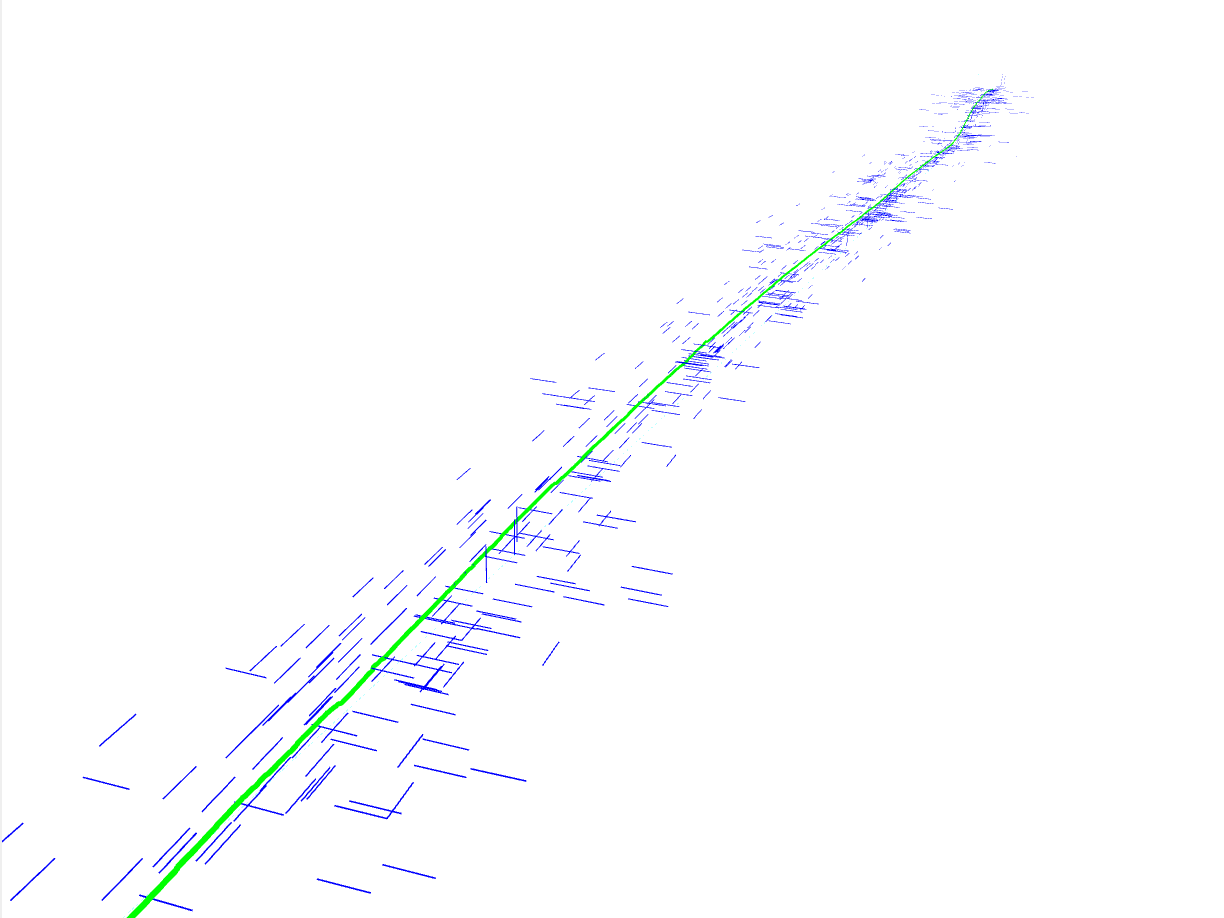}}
    \caption{Line triangulation result in KAIST Urban23 (a) PL-VINS results: triangulated lines are shown in red; 
    (b) PL-VIWO results: triangulated lines are shown in blue.
    }
    \label{fig:Triangulatuon}
\end{figure}

\begin{figure}[!t]
    \centering
    \subfloat[]{\includegraphics[width=0.25\textwidth]{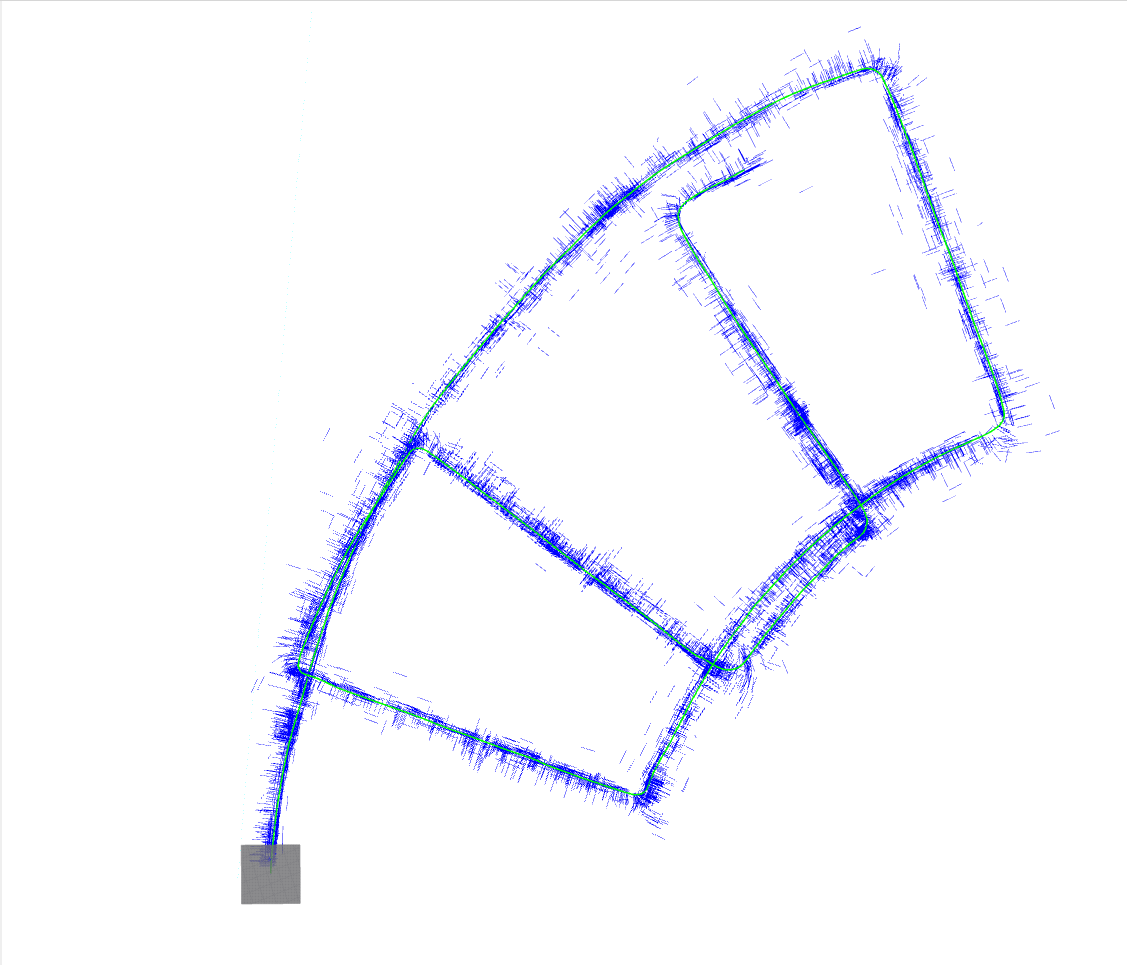}}
    \subfloat[]{\includegraphics[width=0.25\textwidth]{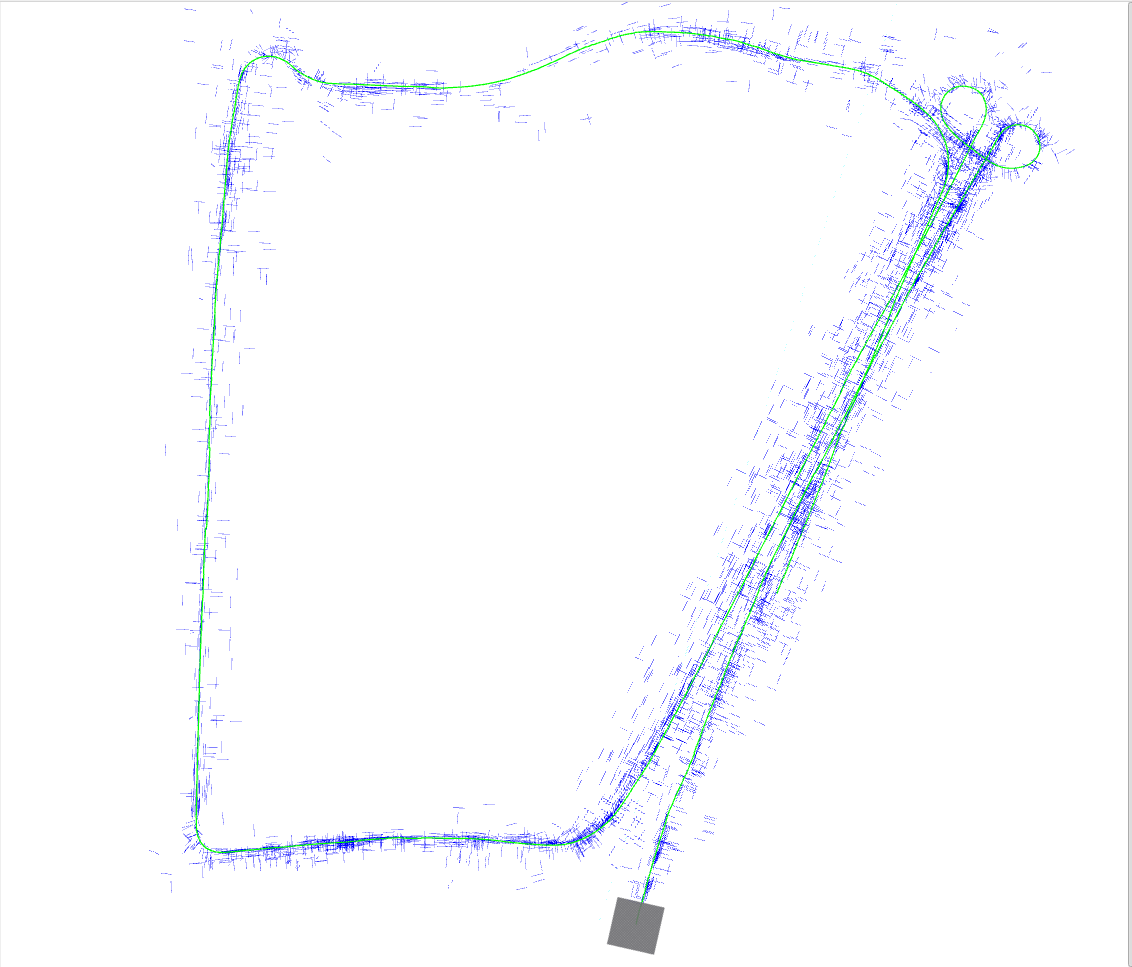}}
    \caption{ Trajectories with lines (a) urban27; 
    (b) urban34.}
    \label{fig:lines}
\end{figure}
To verify the effectiveness of the proposed line feature triangulation methods, we present the triangulation results of PL-VINS and PL-VIWO on highway scenes from KAIST Urban23 in Fig.\ref{fig:Triangulatuon}. PL-VINS triangulation relies solely on the intersection of two planes, and is degenerate for most 2D straight movements, resulting in a limited number of lines successfully triangulated. Ours incorporates the point-line relationship and the number of triangulated lines increases significantly. This indicates that more geometric constraints are utilized in the state update, enhancing accuracy and robustness. Two trajectories with lines in complex urban sequences (urban27 and urban34) are shown in Fig. \ref{fig:lines}. 

\section{CONCLUSIONS}
In this paper, we propose a low-cost monocular VIWO system that is robust and accurate for complex outdoor navigation. To enforce visual constraints, we integrate line features through a novel processing pipeline that leverages the point-line relationship. This approach enables efficient tracking and triangulation of line features, ultimately enhancing system accuracy. Additionally, to improve reliability in dynamic environments, we incorporate MCC to ensure a sufficient number of valid feature points for visual updates. Extensive experiments validate the efficiency and accuracy of our approach. In future work, we plan to extend PL-VIWO to a stereo version and further investigate constraints between points and lines, like point-on-line and parallel lines.

\addtolength{\textheight}{-0cm}   


\end{document}